\documentclass[journal]{IEEEtran}
\usepackage{cite}
\usepackage{amsmath,graphicx}
\usepackage{epstopdf}
\usepackage{xcolor}
\usepackage{amssymb}
\usepackage{pifont}
\usepackage{array}
\usepackage{booktabs}
\usepackage{bm}
\usepackage{multirow}
\usepackage{graphicx}
\usepackage{color}
\usepackage{float}
\usepackage{stfloats}
\usepackage[misc]{ifsym}
\usepackage{stmaryrd}
\usepackage{grffile}
\usepackage{subfig}
\usepackage{makecell}
\usepackage{threeparttable}
\usepackage{enumitem}
\usepackage{enumerate}
\usepackage[colorlinks,linkcolor=red]{hyperref}

\begin{document}
 \UseRawInputEncoding 
\title{DARDet: A Dense Anchor-free Rotated Object Detector in Aerial Images}
\author{Feng Zhang, Xueying Wang, Shilin Zhou, Yingqian Wang
\thanks{This work was partially supported in part by the National Natural Science Foundation of China (Nos. 61903373, 61401474, 61921001).}
\thanks{Feng~Zhang, Xueying~Wang, Shilin~Zhou, Yingqian~Wang are with the College of Electronic Science and Technology, National University of Defense Technology, P. R. China. Emails: \{zhangfeng01, wangxueying, slzhou, wangyingqian16\}@nudt.edu.cn. (Corresponding author: Xueying~Wang)}}

\markboth{Submitted to IEEE Geoscience and Remote Sensing Letters}%
{Shell \MakeLowercase{\textit{et al.}}: Bare Demo of IEEEtran.cls for IEEE Journals}

\maketitle
\begin{abstract}
Rotated object detection in aerial images has received increasing attention for a wide range of applications. However, it is also a challenging task due to the huge variations of scale, rotation, aspect ratio, and densely arranged targets. Most existing methods heavily rely on a large number of pre-defined anchors with different scales, angles, and aspect ratios, and are optimized with a distance loss. Therefore, these methods are sensitive to anchor hyper-parameters and easily suffer from performance degradation caused by boundary discontinuity. To handle this problem, in this paper, we propose a dense anchor-free rotated object detector (DARDet) for rotated object detection in aerial images. Our DARDet directly predicts five parameters of rotated boxes at each foreground pixel of feature maps. We design a new alignment convolution module to extracts aligned features and introduce a PIoU loss for precise and stable regression. Our method achieves state-of-the-art performance on three commonly used aerial objects datasets (i.e., DOTA, HRSC2016, and UCAS-AOD) while keeping high efficiency. Code is available at
 \url{https://github.com/zf020114/DARDet}.
\end{abstract}
\begin{IEEEkeywords}
Rotated object detection, Aerial images, Deep convolution neural networks, Anchor-free detector.
\end{IEEEkeywords}

\section{Introduction}\label{introduction}
\IEEEPARstart{O}{bject} detection in aerial images plays an important role in numerous applications such as intelligent transportation, port management, and urban planning \cite{li2021gated} \cite{ShipSRDet} \cite{MEAD}. However, it is also a challenging task since the objects typically have different scales and aspect ratios \cite{CHPDet}. Moreover, the objects are usually displayed in arbitrary directions and are densely packed. The oriented bounding box (OBB) can compactly enclose the target object and is widely used for aerial image object detection \cite{he2021enhancing}.

Most existing rotated object detection methods are anchor-based frameworks. They introduce an additional angle dimension based on general rectangular bounding box object detection and use simple distance loss. For instance, RoI-Trans \cite{Ding.2018} learns the spatial transformation from horizontal bounding boxes (HBB) to OBB. In S$^{2}$A-Net \cite{s2anet}, an anchor refinement network is proposed to generate high-quality anchors, and only one square anchor is used at each pixel of feature maps. SCRDet \cite{SCRDet} applies an IoU-smooth L1 loss to smooth the loss near the boundary. CSL\cite{CSL} converts angle regression into a classification task to handle the boundary problem. Although these methods have achieved promising performance, these models are sensitive to anchor hyper-parameters or easily suffer from the performance decrease caused by boundary discontinuity issue (see Section \ref{piou loss}).
\begin{figure}
\centering
\includegraphics[width=8.8cm]{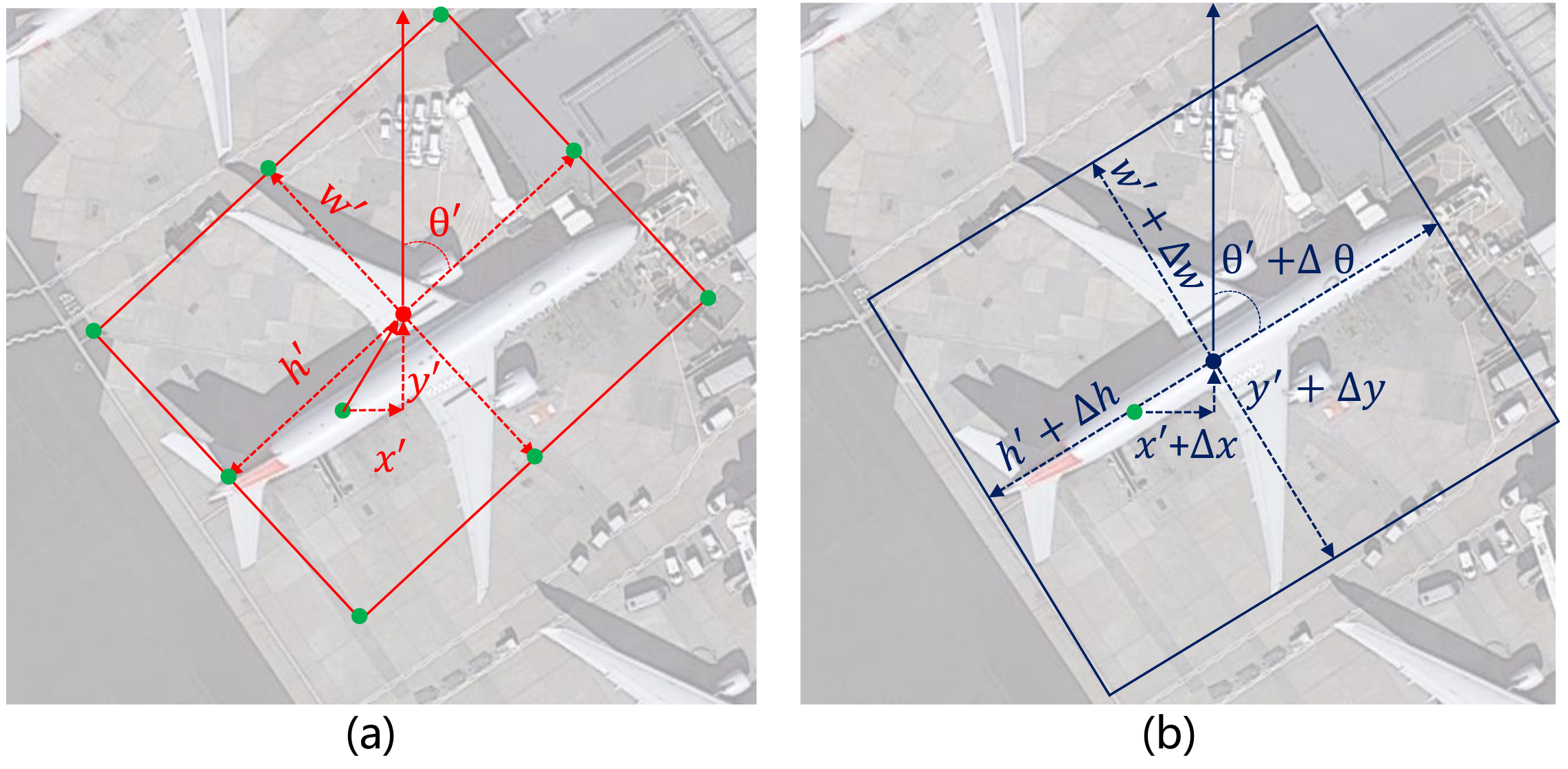}
\caption{DARDet directly predicts a 5D vector $(x^{\prime},y^{\prime},w^{\prime},h^{\prime},\theta^{\prime})$ to encode the OBB at each foreground pixel. (a) shows the initial stage. The green dots illustrate the sampling locations of alignment convolution module. We use these locations as offset field and fed them into deformable convolutions to extract aligned features. (b) shows the refining stage, we predict a 5D offset vector $ (\varDelta x,\varDelta y,\varDelta w,\varDelta h,\varDelta \theta)$ based on the aligned feature. The initially regressed OBB (in red) is then refined into a more accurate one (in blue) by adding the offset.
}\label{Reps}
\end{figure}

Anchor-free detectors can avoid hyperparameters related to anchor boxes by eliminating the predefined anchor boxes. At present, several anchor-free rotated detectors are proposed. For example, BBAVectors \cite{BBAVectors} uses the box boundary-aware vectors to present OBB. PolarDet \cite{PolarDet} represents the oriented objects by using polar coordinate. VCSOP \cite{Orientation-aware} use one subnet to search central points and the remaining three subnets to predict other parameters. However, these methods are all keypoint-based detectors, which encode only one training sample for each annotated OBB and thus result in a long training time. Moreover, these methods usually present rotated boxes in a complex form to solve the boundary discontinuity issue. The features used in these methods are not aligned with the rotated boxes. These drawbacks impede the accuracy of the detectors. In contrast, dense anchor-free prediction frameworks, which pre-define marks on a dense feature map grid and encode multiple training samples for an annotation \cite{sparse}, are more suitable for oriented object detection tasks.

In this paper, we propose a dense anchor-free rotated object detector based on the VarifocalNet \cite{varifocalnet} architecture. VarifocalNet is a dense object detector based on FCOS. It learns IoU-aware classification scores (IACS) that simultaneously represent the classification confidence and localization accuracy. In this paper, we extend the VarifocalNet to rotated object detection. In the regression branch, as shown in Fig.~\ref{Reps}, instead of predicting a 4D vector $(l,t,r,b)$, we directly predict a 5D vector $(x^{\prime},y^{\prime},w^{\prime},h^{\prime},\theta^{\prime})$ encoding the OBB. An efficient alignment convolution module (ACM) is designed to align features with the OBB. We also introduce PIoU loss \cite{piou} to handle the boundary discontinuity problem. 
The main contributions of this paper are summarized as follows.
\begin{itemize}
\item We develop a new dense anchor-free rotated object detection architecture (DARDet), which directly predicts five parameters of OBB at each spatial location. 
\item We design a new efficient alignment convolution module to extract aligned features, which is used to refine coarse OBB and estimate IACS.
\item We introduce PIoU loss to effectively handle the boundary discontinuity problem and achieve state-of-the-art performance on the DOTA, UCAS-AOD, and HRSC2016 datasets with high efficiency.
\end{itemize}

\begin{figure*}
\centering
\includegraphics[width=17.6cm]{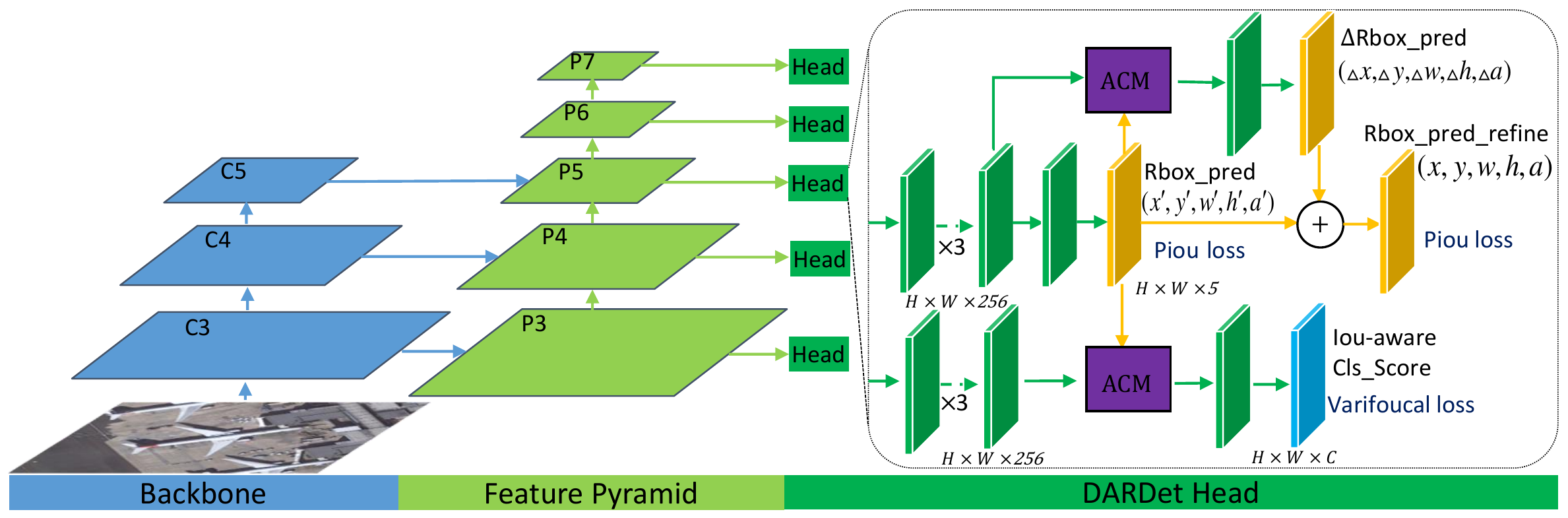}
\caption{An overview of our DARDet. Here, $C3-C5$ denote the feature maps of the backbone and $P3-P7$ denote the feature pyramid. Convolution layers in the same color have the same channels in the DARDet head. The DARDet head contains two sub-networks. The localization subnet consists of two stages. In the initial stage, we direct predict five parameters of OBB and get aligned features by ACM. In the refining stage, we refine the OBB by predicting a 5D offset vector. The other classification sub-network is used to predict the IACS optimized by Varifocal loss.}
\label{framework}
\end{figure*}

\section{Proposed Method}\label{PROPOSED METHOD}

In this section, we first introduce our DARDet, then describe the alignment convolution module. Finally, we introduce the PIoU loss.

\subsection{Overall Pipeline}\label{Pipeline}
As shown in Fig.~\ref{framework}, our DARDet consists of a feature extraction model and DARDet head. The feature extraction model consists of a backbone and FPN. The DARDet head consists of two subnets. The localization subnet takes the feature map from each level of the feature pyramid as its input and applies three $3\times3$ convolution layers to produce a feature map with 256 channels. It performs OBB regression and subsequent refinement in the initial stage and the refining stage, respectively. In the initial stage, the subnet applies a convolution layer to generate a 5D vector $(x^{\prime},y^{\prime},w^{\prime},h^{\prime},\theta^{\prime})$ for each spatial location and extract aligned features by ACM. In the refining stage, the subnet convolves the aligned feature map to produce a 5D offsets vector $ (\varDelta x,\varDelta y,\varDelta w,\varDelta h,\varDelta \theta)$ which is added with the initial OBB location vector to generate the refined OBB $ (x, y, w, h, \theta)$. PIoU loss is used to optimize the OBB. The other classification subnet is used for estimating the IACS, which has a similar architecture with the refining stage of the localization subnet. It outputs IACS, a vector with $C$ (the class number) channels, which represents both the classification confidence and localization accuracy. We used Varifocal loss for training the dense rotated object detector to predict the IACS.
 
\begin{figure}
\centering
\includegraphics[width=8.8cm]{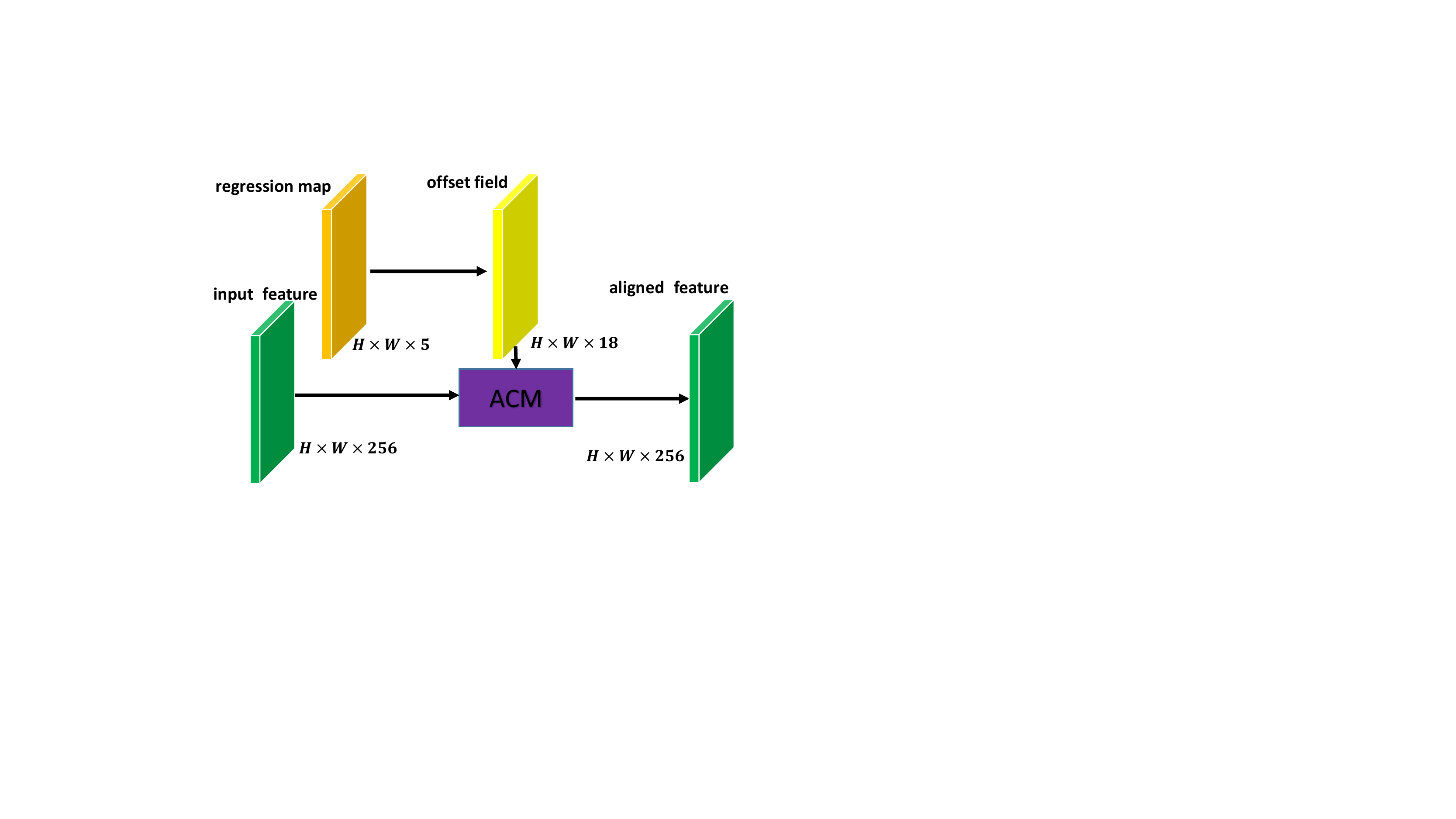}
\caption{Alignment convolution module takes the input feature and the regression map as inputs and
produces aligned features.}\label{ACM_}
\end{figure}

\begin{figure}
\centering
\includegraphics[width=8.8cm]{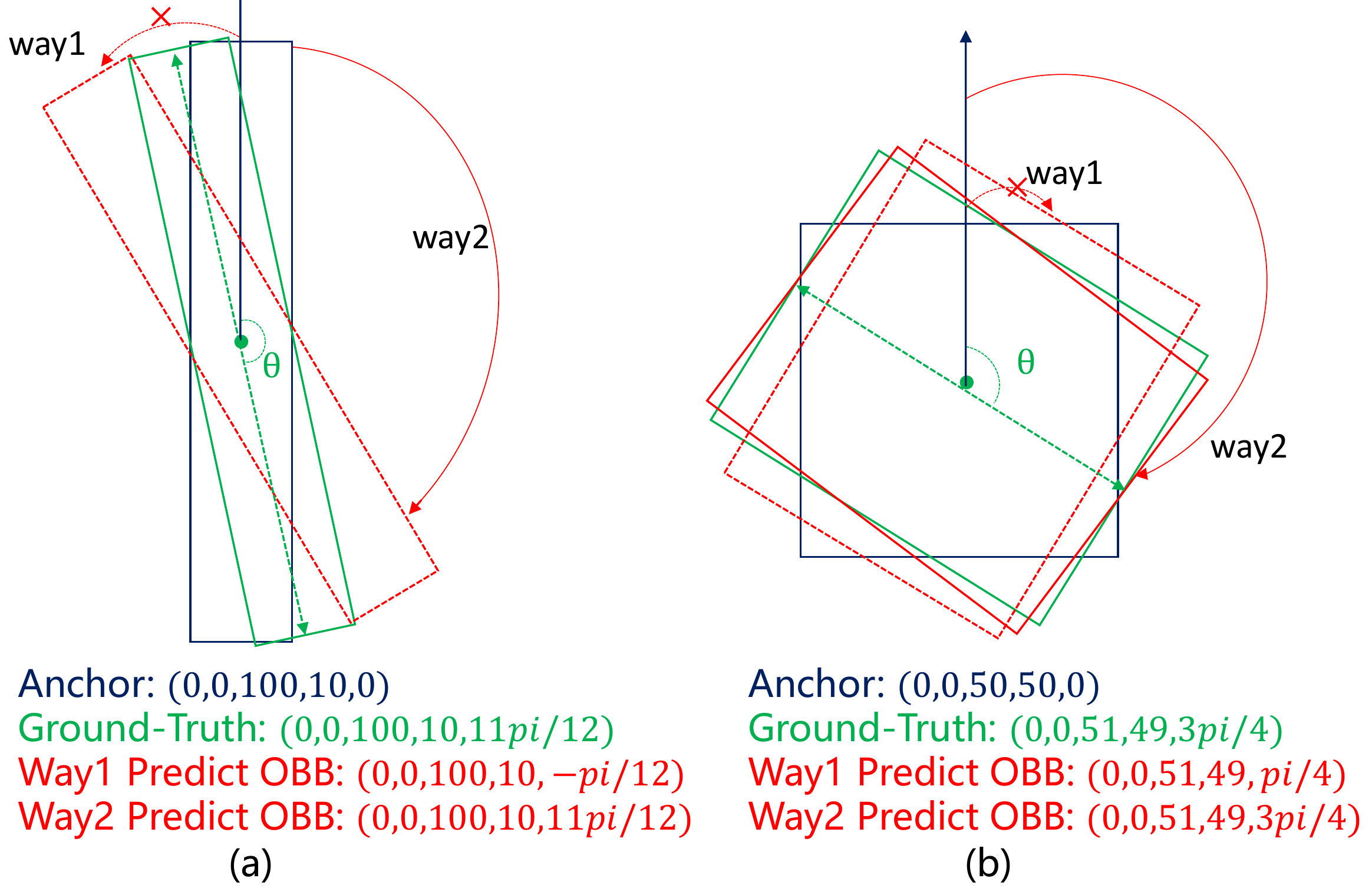}
\caption{An illustration of boundary discontinuity problem. The solid and the dotted arrow indicate the correct and wrong regression way, respectively. (a) illustrates the boundary discontinuity due to the periodicity of angles. (b) shows boundary discontinuity caused by the changeable edges.}\label{Boundary}
\end{figure}

\subsection{Alignment Convolution Module}\label{ACM}
We design a new efficient ACM which can capture the geometry information of an OBB and its nearby contextual information. This information is essential for reducing the misalignment between the predicted OBB and the ground-truth one. This module aligns features with OBB by using the representation with deformable convolution. Specifically, given a sampling location $(i,j)$ on feature map, we first regress an initial OBB vector $(x^{\prime},y^{\prime},w^{\prime},h^{\prime},\theta^{\prime})$. With this initial OBB, we heuristically select 9 sampling points (the 4 vertices, midpoints of the 4 edges of the OBB, and the sampling location). As shown in Fig.~\ref{Reps}(a). the green dots in the figure represent 9 sampling localizations. These 9 localizations are then mapped onto the feature map and features at the projecting points are convolved by the deformable convolution to extract aligned features, as shown in Fig.~\ref{ACM_}. This new module is computationally efficient because these points are manually selected with negligible additional computation cost.

\subsection{PIoU Loss}\label{piou loss}

Most rotation detectors suffer the problem of boundary discontinuity. The boundary discontinuity refers to the sharp loss increase at the boundary due to the periodicity of the angular and exchangeable property of edges. As shown in Fig. \ref{Boundary}, when the angle is near the boundary, it is difficult to judge whether the angle is near $0^{\circ}$ or near $180^{\circ}$, because the definition range of angle is $[0^{\circ}-180^{\circ})$. Similarly, when the OBB is approximately square, it is difficult for the model to distinguish which angle to regress, because the definition of angle is the angle between the long edge and the y-axis. When the judgment is wrong, the distance loss will be very large. 

\begin{figure}
\centering
\includegraphics[width=8.8cm]{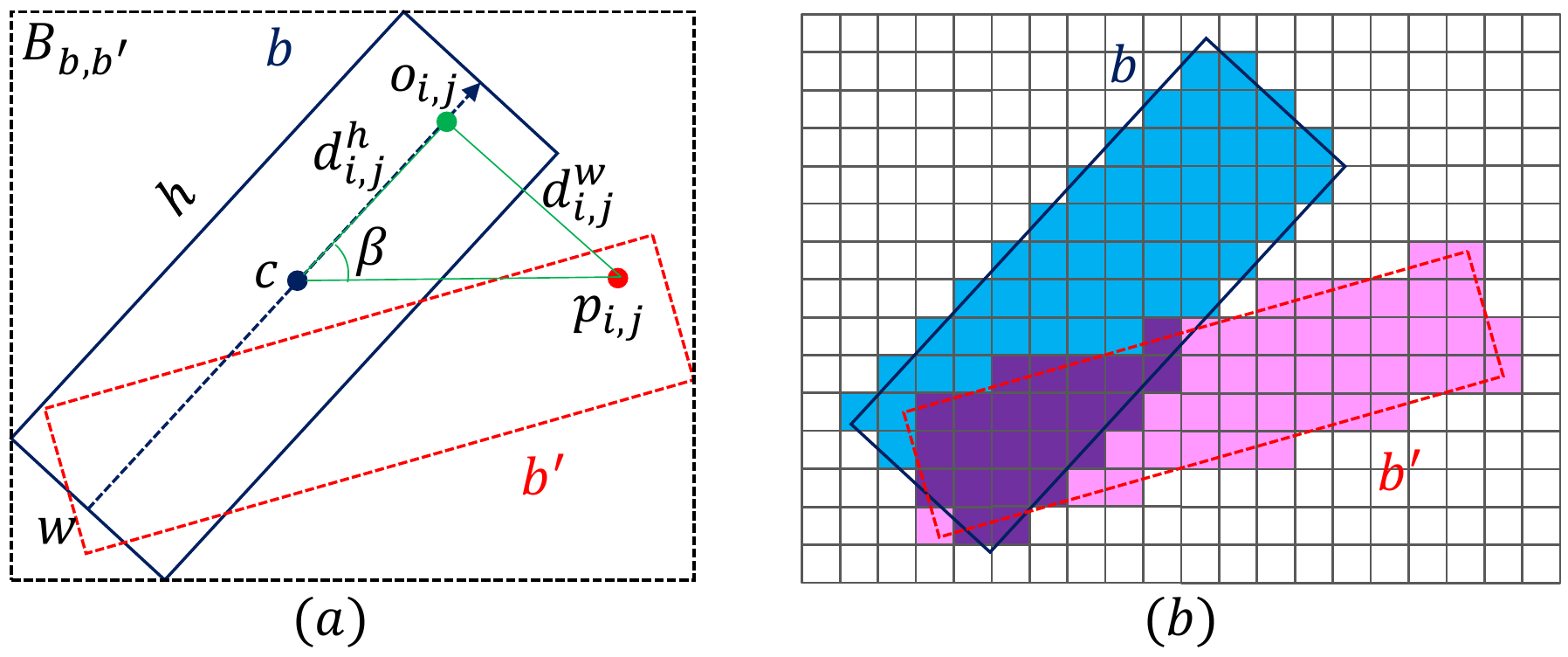}
\caption{Schematic diagram of calculating PIoU. (a) Components involved in determining the relative position (inside or outside) between a pixel $p_{i,j}$ (red point) and an OBB $b$ (blue rotated box). $O_{i,j}$ (green point) is the intersection point between the mid-perpendicular line and its orthogonal line to $p_{i,j}$. OBB center $c$ (blue point), $p_{i,j}$ and $o_{i,j}$ form a triangle. The length of the two right-angled edges of the triangle are denoted by $d^{h}_{i,j}$, $d^{w}_{i,j}$. We use a binary function to judge the relative location. (b) Statistics of all pixels to calculate PIoU. The intersection area is in purple, and the area made up of yellow, blue and purple is the union region.}\label{loss}
\end{figure}

Pixels-IoU loss (PIoU) can solve boundary discontinuity by jointly correlating the five parameters of OBB to check the position (inside or outside the intersection) and contribution of each pixel. The intersection area is calculated by the number of interior pixels. As shown in Fig.~\ref{loss} (a), for a given OBB $b$ (blue rotated box encoded by $ (x, y, w, h, \theta)$) and a pixel $p_{i,j}$ in image, to judge the relative location between $p_{i,j}$ and $b$, we define the binary function as follows:

\begin{equation}\label{binary}
\delta\left(p_{i, j} \mid b\right)=\left\{\begin{array}{ll}1, & d_{i, j}^{w} \leq \frac{w}{2}, d_{i, j}^{h} \leq \frac{h}{2} \\ 0, & \text { otherwise }\end{array}\right.,
\end{equation}
where $d^h_{i,j}$ and $d^w_{i,j}$ denote the distance along horizontal and vertical direction, respectively. Since Eq. \ref{binary} is not continuous and differentiable, we approximate this binary function by multiplying the following two kernels:
\begin{equation}
F\left(\boldsymbol{p}_{i, j} \mid \boldsymbol{b}\right)=K\left(d_{i, j}^{w}, w\right) K\left(d_{i, j}^{h}, h\right).
\end{equation}
 
The kernel function $K(d, s)$ is defined by
\begin{equation}
K(d, s)=1-\frac{1}{1+e^{-k(d-s)}}.
\end{equation}

The intersection area $S_{\boldsymbol{b} \cap \boldsymbol{b}^{\prime}}$ and union area $S_{\boldsymbol{b} \cup \boldsymbol{b}^{\prime}}$ between $b$ and $b'$ are approximated by:
\begin{equation}
S_{\boldsymbol{b} \cap \boldsymbol{b}^{\prime}} \approx \sum_{\boldsymbol{p}_{i, j} \in B_{\boldsymbol{b}, \boldsymbol{b}^{\prime}}} F\left(\boldsymbol{p}_{i, j} \mid \boldsymbol{b}\right) F\left(\boldsymbol{p}_{i, j} \mid \boldsymbol{b}^{\prime}\right),
\end{equation}
\begin{equation}
S_{\boldsymbol{b} \cup \boldsymbol{b}^{\prime}} \approx w \times h+w^{\prime} \times h^{\prime}-S_{\boldsymbol{b} \cap \boldsymbol{b}^{\prime}}.
\end{equation}

Then PIoU is computed as:
\begin{equation}
PIoU\left(b, b^{\prime}\right)=\frac{S_{b \cap b^{\prime}}}{S_{b \cup b^{\prime}}}.
\end{equation}

The PIoU loss of the two regression ways in Fig.~\ref{Boundary} is almost equal and the loss is continuous under the boundary condition. Therefore, it can well handle the boundary discontinuity problem. 

\section{Experiments}\label{Experiments}

\subsection{Datasets}
We evaluate our method on the DOTA, HRSC2016, and UCAS-AOD datasets. 

\subsubsection{DOTA} DOTA is a large dataset for oriented object detection. It contains 2806 images and 15 common object categories. In the ablation study, we use the training set for training and the validation set for evaluation. We cropped the images into $1024\times 1024$ patches with a stride of 824 and performed random flipping for data augmentation. For comparison with other methods, all images are cropped with a stride of 512, and we only use a single scale for training and testing. 


\subsubsection{HRSC2016} The HRSC2016 dataset is a challenging dataset for ship detection in aerial images. The training, validation, and test sets include 436, 181, and 444 images, respectively.  We used the training and validation sets for training and evaluated the performance on the test set. All images were cropped to patches of size $1024\times1024$. 

\subsubsection{UCAS-AOD} The UCAS-AOD dataset contains 1510 aerial images and 14596 instances of two categories including plane and car. We randomly sampled 1132 images for training and 378 images for testing. All images were cropped into patches of size $672 \times 672$.

\begin{table}
\centering
\footnotesize
\caption{Ablation study for DARDet on the DOTA dataset. We use the training set for training and validation set for evaluation.
}\label{Ablation}
\begin{tabular}{c|cccc}
\Xhline{1pt}
                             &\multicolumn{3}{c}{Different Settings of DARDet} \\
\Xhline{1pt}
Alignment Convolution Module  &            &$\checkmark$&$\checkmark$\\
PIoU loss                     &            &            &$\checkmark$\\
\hline 
mAP                           &  63.19    & 66.98      & 72.44      \\
\Xhline{1pt}
\end{tabular}
\end{table}

\subsection{Implementation Details}

In the ablation study, DARDet was trained for 12 epochs with a batch size of 6, and ResNet50 was used as the backbone. For performance comparison, we expanded the receptive field of our detector by replacing ordinary convolutions with deformable convolutions at the last stage of the backbone and trained the models for 24 epochs. In all experiments, SGD optimizer was adopted with an initial learning rate of 0.01 and the learning rate was divided by 10 at each decay step. The momentum and weight decay were 0.9 and 0.0001, respectively. The result was measured on NVIDIA RTX 2080Ti GPU with a batch size of 1 and $1024\times1024$ image size with VOC2007 metrics.
\subsection{Ablation Study}

\subsubsection{Modified VarifocalNet as baseline} We extend the VarifocalNet architecture to the oriented object detection task by directly predicting a 5D vector to encode the OBB and using smooth L1 loss in the regression branch. As shown in Table \ref{Ablation}, Modified VarifocalNet achieves an mAP of 63.19$\%$, which demonstrates that our baseline achieves competitive performance.

\subsubsection{Effectiveness of Alignment Convolution Module} We use the alignment convolution module to extract aligned features and then simultaneously estimate the IACS and refine the OBB. As shown in Table \ref{Ablation}, compared with the baseline, our method improves mAP by 3.8$\%$ to 66.98$\%$.

\subsubsection{Effectiveness of PIoU loss} When we replace the smooth L1 loss with PIoU loss, the overall mAP is improved from 66.98$\%$ to 72.44$\%$. It demonstrates the effectiveness of the PIoU loss. 

\begin{figure}
\centering
\includegraphics[width=8.8cm]{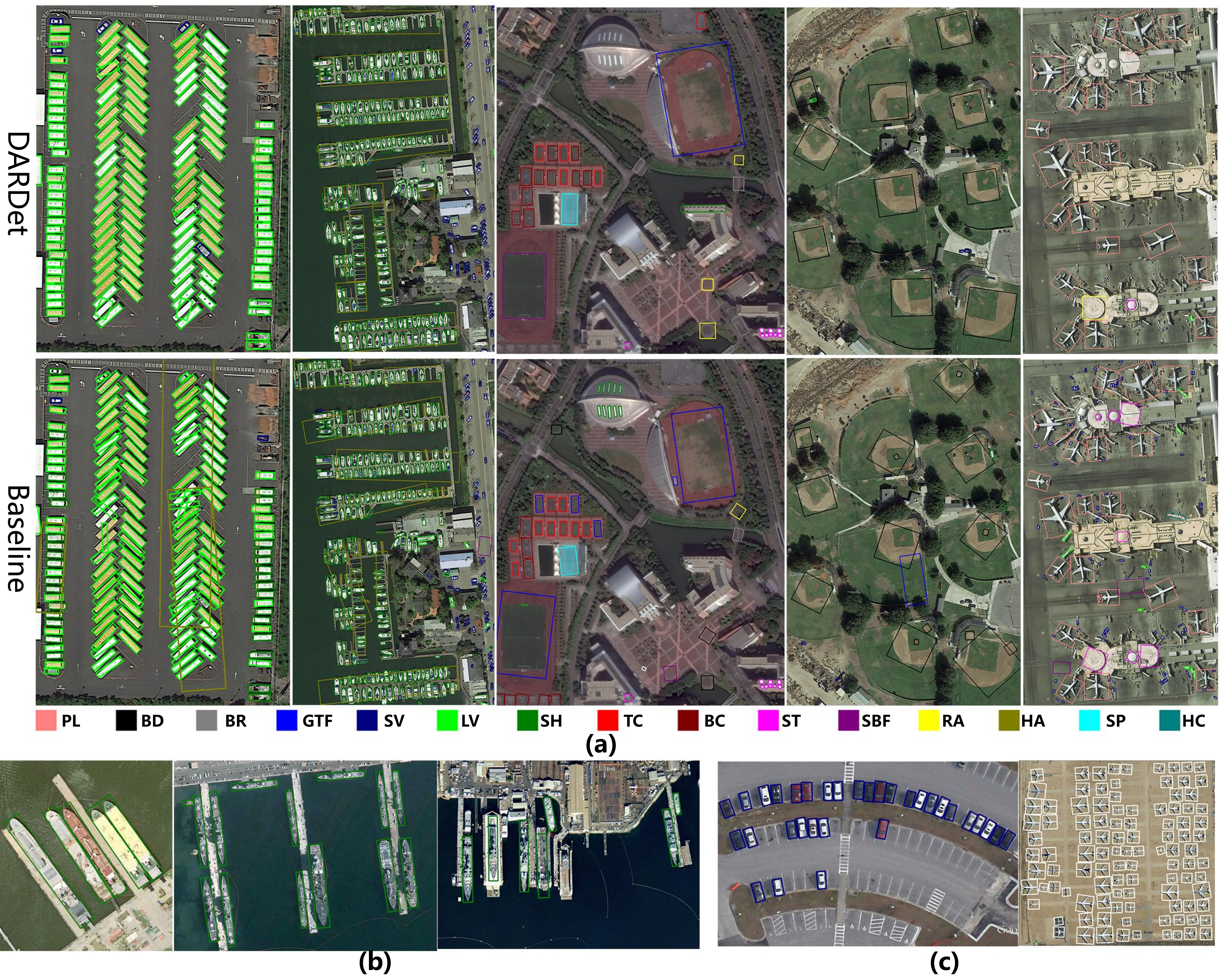}
\caption{Some detection results on three commonly used aerial objects datasets. (a) is qualitative comparisons between the proposed DARDet and the baseline method on the DOTA dataset. (b) and (c) are the sample results achieved by DARDet on the HRSC2016 and UCAS-AOD datasets, respectively. Our method achieves promising results when detecting dense, rotated, clutter, and high aspect ratio objects.}\label{example result}
\end{figure}

Qualitative detection results of the baseline method and our DARDet are visualized in Fig.~\ref{example result} (a). As shown in the first three columns of the figure, almost all the objects are detected correctly. These results suggest that our method can handle challenging situations even with a high aspect ratio or very cluttered scenes. In the last two columns, the main objects in the images are baseball diamonds and planes. Some objects are almost equal in length and width, so the use of distance loss will encounter boundary problems. Specifically, it is difficult for the model to distinguish which side is longer. To reduce loss, the model regresses the angle to the intermediate value of the two possible correct angles. Therefore, we can see that almost all the baseball diamonds and some planes in the baseline results deviate from the ground truth by $45^{\circ}$. However, our method can solve the boundary problems effectively and obtains accurate results.

\begin{table*}
    \scriptsize
\centering
\caption{Comparisons with state-of-art methods for the DOTA OBB task. Rot indicates rotated augmentation in training. The results in red and blue indicate the best and second-best results of each column. Only sigle-scale training and testing is used in experiments.} \label{dota_result} %
\begin{tabular}{p{2.0cm}<{\centering} p{1.1cm}<{\centering} p{0.33cm}<{\centering} p{0.33cm} p{0.33cm} p{0.33cm} p{0.33cm} p{0.33cm} p{0.33cm} p{0.33cm} p{0.33cm} p{0.33cm} p{0.33cm} p{0.33cm} p{0.33cm} p{0.33cm} p{0.33cm} p{0.4cm}  p{0.4cm} c c <{\centering}}
\Xhline{1pt}  
Method                        &Backbone &Rot & PL & BD & BR & GTF & SV & LV & SH & TC & BC & ST & SBF & RA & HA & SP & HC  & mAP &FPS \\
\Xhline{1pt}
ROI-Trans\cite{Ding.2018}       &ResNet101 & &88.64&78.52&43.44&\textcolor{blue}{75.92}&68.81&73.68&83.59&90.74&77.27&81.46&58.39&53.54&62.83&58.93&47.67&69.56&9.4\\
SCRDet\cite{SCRDet}             &ResNet50&   &89.48&79.83&50.43&62.51&76.72&74.31&85.42&90.86&84.19&82.51&42.84&60.15&74.18&70.62&51.81&71.72&4.4\\
Redet\cite{redet}               &ReResNet50 &&88.79&82.64&53.97&74.00&78.13&\textcolor{red}{84.06}&88.04&90.89&\textcolor{red}{87.78}&85.75&61.76&60.39&75.96&68.07&63.59&76.25&9.2\\
R$^3$Det\cite{R3d}              &ResNet101 & &88.76&83.09&50.91&67.27&76.23&80.39&86.72&90.78&84.68&83.24&61.98&61.35&66.91&70.63&53.94&73.79&6.9\\
RSDet \cite{rsdet}              &ResNet152&  &\textcolor{red}{90.2}&83.5&53.6&70.1&64.6&79.4&67.3&91.0&88.3&82.5&64.1 &\textcolor{red}{68.7} &62.8 &69.5&66.9&73.5&7.1\\
R$^{3}$Det-KLD\cite{KLD}        &ResNet50&   &88.90&84.17&\textcolor{blue}{55.80}&69.35&78.72&84.08&87.00&89.75&84.23&85.73&\textcolor{red}{64.74}&61.80&76.72&78.49&\textcolor{red}{70.89}&77.36&\_\\
R$^{3}$Det-GWD\cite{GWD}        &ResNet50&   &88.82&82.94&55.63&72.75&78.52&83.10&87.46&90.21&\textcolor{red}{86.36}&85.44&\textcolor{blue}{64.70}&61.41&73.46&76.94&57.38&76.34&\_\\
CSL\cite{CSL}                   &ResNet50&   &89.09&79.22&45.78&67.94&66.58&56.29&71.55&90.80&80.50&76.61&58.53&61.01&59.92&69.51&57.59&68.73&5.1\\
S$^{2}$A-Net\cite{s2anet}       &ResNet50 &  &89.11&82.84&48.37&71.11&78.11&78.39&87.25&90.83&84.90&85.64&60.36&62.60&65.26&69.13&57.94&74.12&\textcolor{red}{13.5}\\
O$^{2}$-DNet\cite{middlelines}&Hourglass104& &89.31&82.14&47.33&61.21&71.32&74.03&78.62&90.76&82.23&81.36&60.93&60.17&58.21&66.98&64.03&71.04&\_\\
BBAVectors \cite{BBAVectors}    &ResNet101&  &88.35&79.96&50.69&62.18&78.43&78.98&87.94&90.85&83.58&84.35&54.13&60.24&65.22&64.28&55.70&72.32&6.9\\
PolarDet\cite{PolarDet}         &ResNet50 &  &\textcolor{blue}{89.73}&\textcolor{red}{87.05}&45.30&63.32&78.44&76.65&87.13&90.79&80.58&85.89&60.97&\textcolor{blue}{67.94}&68.20&74.63&68.67&75.02&\_\\
\hline
DARDet (ours)                  &ResNet50 &  &88.89&\textcolor{blue}{84.31}&55.32&75.49&\textcolor{blue}{80.33}&81.69&\textcolor{blue}{88.24}&\textcolor{blue}{90.88}&83.62&\textcolor{blue}{87.46}&59.85&65.60&\textcolor{blue}{76.86}&\textcolor{blue}{80.46}&65.17&\textcolor{blue}{77.61}&\textcolor{blue}{12.6}\\
DARDet (ours)      &ResNet50 &$\checkmark$&89.08&84.30&\textcolor{red}{56.64}&\textcolor{red}{77.83}&\textcolor{red}{81.10}&\textcolor{blue}{83.39}&\textcolor{red}{88.46}&\textcolor{red}{90.88}&85.44&\textcolor{red}{87.56}&62.77&66.23&\textcolor{red}{77.97}&\textcolor{red}{82.03}&\textcolor{blue}{67.40}&\textcolor{red}{78.74}&\textcolor{red}{12.6}\\

\Xhline{1pt}
\end{tabular}
\end{table*}

\subsection{Comparison with the SOTA Methods}
\subsubsection{Results on DOTA}
We compare our DARDet with other state-of-the-art detectors for the OBB task. As shown in Table \ref{dota_result}, our single-scale DARDet with ResNet50 achieves an mAP of 77.61$\%$, which outperforms all single-scale models without bells and whistles. Our DARDet achieves state-of-the-art detection accuracies (78.74$\%$ in mAP) with rotated augmentation, as compared with other single-scale methods. In terms of the inference speed, our DARDet is much faster than the other methods except S$^{2}$A-Net, but the mAP of our method was higher than S$^{2}$A-Net by 3.5$\%$.
\subsubsection{Results on HRSC2016}
The HRSC2016 dataset contains lots of thin and long ships with arbitrary orientations. The comparative results on the HRSC2016 dataset are shown in Table \ref{HRSC-result}. Our method achieves the best performance (with an accuracy of $90.37\%$.)
\begin{table}
    \footnotesize
\centering
\caption{Detection accuracy on the HRSC2016 dataset.}\label{HRSC-result}
\begin{tabular}{c|c|c}
\Xhline{1pt}
Method                         & Backbone      & mAP$_{0.5}$ \\
\Xhline{1pt}
ROI-trans\cite{Ding.2018}      & Resnet101     & 86.20 \\
R$^3$Det \cite{R3d}            & Resnet101     & 89.33 \\
RSDet \cite{rsdet}             & ResNet152     & 89.43 \\
BBAVectors\cite{BBAVectors}    & Resnet101     & 88.60 \\
S$^{2}$A-Net \cite{s2anet}     & Resnet101     & 90.17 \\
FR-Est \cite{pointbased}       & Resnet101     & 89.7 \\
\hline     
DARDet(ours)                   & Resnet50      &\textbf{90.37} \\
\Xhline{1pt}
\end{tabular}
\end{table}

\subsubsection{Results on UCAS-AOD}
Our method also achieves the best performance on the UCAS-AOD dataset, with an mAP of $90.37\%$. Table \ref{AOD} shows the comparative results.

\begin{table}
    \footnotesize
\centering
\caption{Detection accuracy on the UCAS-AOD dataset.}\label{AOD}
\begin{tabular}{c|c|c|c|c}
\Xhline{1pt}
Method                         & Backbone  & car  & airplane &mAP$_{0.5}$\\
\Xhline{1pt}
FR-O\cite{dota}                & Resnet101 &86.87  &89.86    &88.36 \\
ROI-trans\cite{Ding.2018}      & Resnet101 &87.99  &89.90    &88.95 \\
FPN-CSL\cite{CSL}              & Resnet101 &88.09  &90.38    &89.23 \\
R$^3$Det-DCL\cite{DCL}         & Resnet101 &88.15  &90.57    &89.36     \\
DAL\cite{DAL}                  & Resnet101 &89.25  &90.49    &89.87 \\
\hline
DARDet(ours)                   & Resnet50  &89.97  &90.76    &\textbf{90.37}\\
\Xhline{1pt}
\end{tabular}
\end{table} 

\section{Conclusion}\label{Conclusion}
In this paper, we proposed a new dense anchor-free detection framework to detect rotated objects in aerial images. Our method detects objects by directly predicting a 5D vector at each foreground pixel. Moreover, an ACM is designed for extracted aligned convolutional features, and PIoU loss is introduced to optimize the OBB. Experimental results on the DOTA, HRSC2016 and UCAS-AOD datasets demonstrate that our method achieves state-of-the-art performance with high efficiency as compared to other detectors.

\bibliographystyle{IEEEtran}
\bibliography{TGRS}

\begin{thebibliography}{10}
\providecommand{\url}[1]{#1}
\csname url@samestyle\endcsname
\providecommand{\newblock}{\relax}
\providecommand{\bibinfo}[2]{#2}
\providecommand{\BIBentrySTDinterwordspacing}{\spaceskip=0pt\relax}
\providecommand{\BIBentryALTinterwordstretchfactor}{4}
\providecommand{\BIBentryALTinterwordspacing}{\spaceskip=\fontdimen2\font plus
\BIBentryALTinterwordstretchfactor\fontdimen3\font minus
  \fontdimen4\font\relax}
\providecommand{\BIBforeignlanguage}[2]{{%
\expandafter\ifx\csname l@#1\endcsname\relax
\typeout{** WARNING: IEEEtran.bst: No hyphenation pattern has been}%
\typeout{** loaded for the language `#1'. Using the pattern for}%
\typeout{** the default language instead.}%
\else
\language=\csname l@#1\endcsname
\fi
#2}}
\providecommand{\BIBdecl}{\relax}
\BIBdecl

\bibitem{li2021gated}
B.~Li, Y.~Guo, J.~Yang, L.~Wang, Y.~Wang, and W.~An, ``Gated recurrent
  multiattention network for vhr remote sensing image classification,''
  \emph{IEEE Transactions on Geoscience and Remote Sensing}, 2021.

\bibitem{ShipSRDet}
S.~He, H.~Zou, Y.~Wang, R.~Li, and F.~Cheng, ``Shipsrdet: An end-to-end remote
  sensing ship detector using super-resolved feature representation,'' in
  \emph{IGARSS}, July 2021.

\bibitem{MEAD}
Z.~He, Z.~Ren, X.~Yang, Y.~Yang, and W.~Zhang, ``Mead: a mask-guided
  anchor-free detector for oriented aerial object detection,'' \emph{Applied
  Intelligence}, pp. 1--16, 2021.

\bibitem{CHPDet}
F.~Zhang, X.~Wang, S.~Zhou, Y.~Wang, and Y.~Hou, ``Arbitrary-oriented ship
  detection through center-head point extraction,'' 2021.

\bibitem{he2021enhancing}
S.~He, H.~Zou, Y.~Wang, R.~Li, F.~Cheng, X.~Cao, and M.~Li, ``Enhancing
  mid–low-resolution ship detection with high-resolution feature
  distillation,'' \emph{IEEE Geoscience and Remote Sensing Letters}, 2021.

\bibitem{Ding.2018}
J.~Ding, N.~Xue, Y.~Long, G.-S. Xia, and Q.~Lu, ``Learning roi transformer for
  oriented object detection in aerial images,'' in \emph{CVPR}, June 2019.

\bibitem{s2anet}
J.~Han, J.~Ding, J.~Li, and G.-S. Xia, ``Align deep features for oriented
  object detection,'' \emph{IEEE Transactions on Geoscience and Remote
  Sensing}, 2021.

\bibitem{SCRDet}
X.~Yang, J.~Yang, J.~Yan, Y.~Zhang, T.~Zhang, Z.~Guo, X.~Sun, and K.~Fu,
  ``Scrdet: Towards more robust detection for small, cluttered and rotated
  objects,'' in \emph{ICCV 2019}.\hskip 1em plus 0.5em minus 0.4em\relax
  {IEEE}, 2019, pp. 8231--8240.

\bibitem{CSL}
X.~Yang and J.~Yan, ``Arbitrary-oriented object detection with circular smooth
  label,'' in \emph{ECCV}.\hskip 1em plus 0.5em minus 0.4em\relax Springer,
  2020, pp. 677--694.

\bibitem{BBAVectors}
J.~Yi, P.~Wu, B.~Liu, Q.~Huang, H.~Qu, and D.~Metaxas, ``Oriented object
  detection in aerial images with box boundary-aware vectors,'' in \emph{WACV},
  2021, pp. 2150--2159.

\bibitem{PolarDet}
P.~Zhao, Z.~Qu, Y.~Bu, W.~Tan, and Q.~Guan, ``Polardet: A fast, more precise
  detector for rotated target in aerial images,'' \emph{International Journal
  of Remote Sensing}, vol.~42, no.~15, pp. 5821--5851, 2021.

\bibitem{Orientation-aware}
F.~Shi, T.~Zhang, and T.~Zhang, ``Orientation-aware vehicle detection in aerial
  images via an anchor-free object detection approach,'' \emph{IEEE
  Transactions on Geoscience and Remote Sensing}, vol.~59, no.~6, pp.
  5221--5233, 2020.

\bibitem{sparse}
P.~Sun, R.~Zhang, Y.~Jiang, T.~Kong, C.~Xu, W.~Zhan, M.~Tomizuka, L.~Li,
  Z.~Yuan, C.~Wang \emph{et~al.}, ``Sparse r-cnn: End-to-end object detection
  with learnable proposals,'' in \emph{CVPR}, 2021, pp. 14\,454--14\,463.

\bibitem{varifocalnet}
H.~Zhang, Y.~Wang, F.~Dayoub, and N.~Sunderhauf, ``Varifocalnet: An iou-aware
  dense object detector,'' in \emph{CVPR}, 2021, pp. 8514--8523.

\bibitem{piou}
Z.~Chen, K.~Chen, W.~Lin, J.~See, H.~Yu, Y.~Ke, and C.~Yang, ``Piou loss:
  Towards accurate oriented object detection in complex environments,'' in
  \emph{ECCVn}.\hskip 1em plus 0.5em minus 0.4em\relax Springer, 2020, pp.
  195--211.

\bibitem{redet}
J.~Han, J.~Ding, N.~Xue, and G.-S. Xia, ``Redet: A rotation-equivariant
  detector for aerial object detection,'' in \emph{CVPR}, 2021, pp. 2786--2795.

\bibitem{R3d}
X.~Yang, J.~Yan, Z.~Feng, and T.~He, ``R3det: Refined single-stage detector
  with feature refinement for rotating object,'' in \emph{AAAI}, vol.~35,
  no.~4, 2021, pp. 3163--3171.

\bibitem{rsdet}
W.~Qian, X.~Yang, S.~Peng, Y.~Guo, and C.~Yan, ``Learning modulated loss for
  rotated object detection,'' 2019.

\bibitem{KLD}
X.~Yang, X.~Yang, J.~Yang, Q.~Ming, W.~Wang, Q.~Tian, and J.~Yan, ``Learning
  high-precision bounding box for rotated object detection via kullback-leibler
  divergence,'' 2021.

\bibitem{GWD}
X.~Yang, J.~Yan, Q.~Ming, W.~Wang, X.~Zhang, and Q.~Tian, ``Rethinking rotated
  object detection with gaussian wasserstein distance loss,'' 2021.

\bibitem{middlelines}
H.~Wei, Y.~Zhang, Z.~Chang, H.~Li, H.~Wang, and X.~Sun, ``Oriented objects as
  pairs of middle lines,'' \emph{ISPRS Journal of Photogrammetry and Remote
  Sensing}, vol. 169, pp. 268--279, 2020.

\bibitem{pointbased}
K.~Fu, Z.~Chang, Y.~Zhang, and X.~Sun, ``Point-based estimator for
  arbitrary-oriented object detection in aerial images,'' \emph{IEEE
  Transactions on Geoscience and Remote Sensing}, 2020.

\bibitem{dota}
G.~Xia, X.~Bai, J.~Ding, Z.~Zhu, S.~J. Belongie, J.~Luo, M.~Datcu, M.~Pelillo,
  and L.~Zhang, ``{DOTA:} {A} large-scale dataset for object detection in
  aerial images,'' in \emph{CVPR}, 2018, pp. 3974--3983.

\bibitem{DCL}
X.~Yang, L.~Hou, Y.~Zhou, W.~Wang, and J.~Yan, ``Dense label encoding for
  boundary discontinuity free rotation detection,'' in \emph{CVPR}, 2021, pp.
  15\,819--15\,829.

\bibitem{DAL}
Q.~Ming, Z.~Zhou, L.~Miao, H.~Zhang, and L.~Li, ``Dynamic anchor learning for
  arbitrary-oriented object detection,'' in \emph{AAAI}, vol.~35, no.~3, 2021,
  pp. 2355--2363.

\end{thebibliography}
\end{document}